\documentclass{bmvc2k}

\usepackage{graphicx}
\usepackage{tikz}
\usepackage{comment}
\usepackage{amsmath,amssymb} 
\usepackage{graphicx}
\usepackage{booktabs}
\usepackage{makecell}
\usepackage{multirow}
\usepackage{multicol}
\usepackage{colortbl}
\usepackage{xcolor}
\usepackage{xspace}
\usepackage{subcaption}
\usepackage[accsupp]{axessibility}  
\newcommand{\ie}{\textit{i}.\textit{e}.,\xspace}
\newcommand{\eg}{\textit{e}.\textit{g}.,\xspace}

\title{CNeRV: Content-adaptive Neural Representation for Visual Data}

\addauthor{Hao Chen$^\text{1}$}{chenh@umd.edu}{1}
\addauthor{Matthew Gwilliam$^\text{1}$}{mattgwilliamjr@gmail.com}{1}
\addauthor{Bo He$^\text{1}$}{bohe@umd.edu}{1}
\addauthor{Ser-Nam Lim}{sernam@gmail.com}{}
\addauthor{Abhinav Shrivastava$^\text{1}$}{abhinav@cs.umd.edu}{1}

\addinstitution{
$^\text{1}$University of Maryland\\
College Park, MD, USA
}

\runninghead{Chen, Gwilliam, He, Lim, Shrivastava}{CNeRV}

\begin{document}

\maketitle

\begin{abstract}
Compression and reconstruction of visual data have been widely studied in the computer vision community. 
More recently, some have used deep learning to improve or refine existing pipelines, while others have proposed end-to-end approaches, including autoencoders and implicit neural representations, such as SIREN and NeRV. 
In this work, we propose \textbf{Ne}ural \textbf{V}isual \textbf{R}epresentation with \textbf{C}ontent-adaptive Embedding (CNeRV), which combines the generalizability of autoencoders with the simplicity and compactness of implicit representation.
We introduce a novel content-adaptive embedding that is unified, concise, and internally (within-video) generalizable, that compliments a powerful decoder with a single-layer encoder.
We match the performance of NeRV, a state-of-the-art implicit neural representation, on the reconstruction task for frames seen during training while far surpassing for frames that are skipped during training (unseen images). 
To achieve similar reconstruction quality on unseen images, NeRV needs  $\textbf{120}\times$ more time to overfit per-frame due to its lack of internal generalization. 
With the same latent code length and similar model size, CNeRV outperforms autoencoders on reconstruction of both seen and unseen images.
We also show promising results for visual data compression.
\end{abstract}

\section{Introduction}

\begin{figure}[t!]
    \centering
    \includegraphics[width=.7\linewidth]{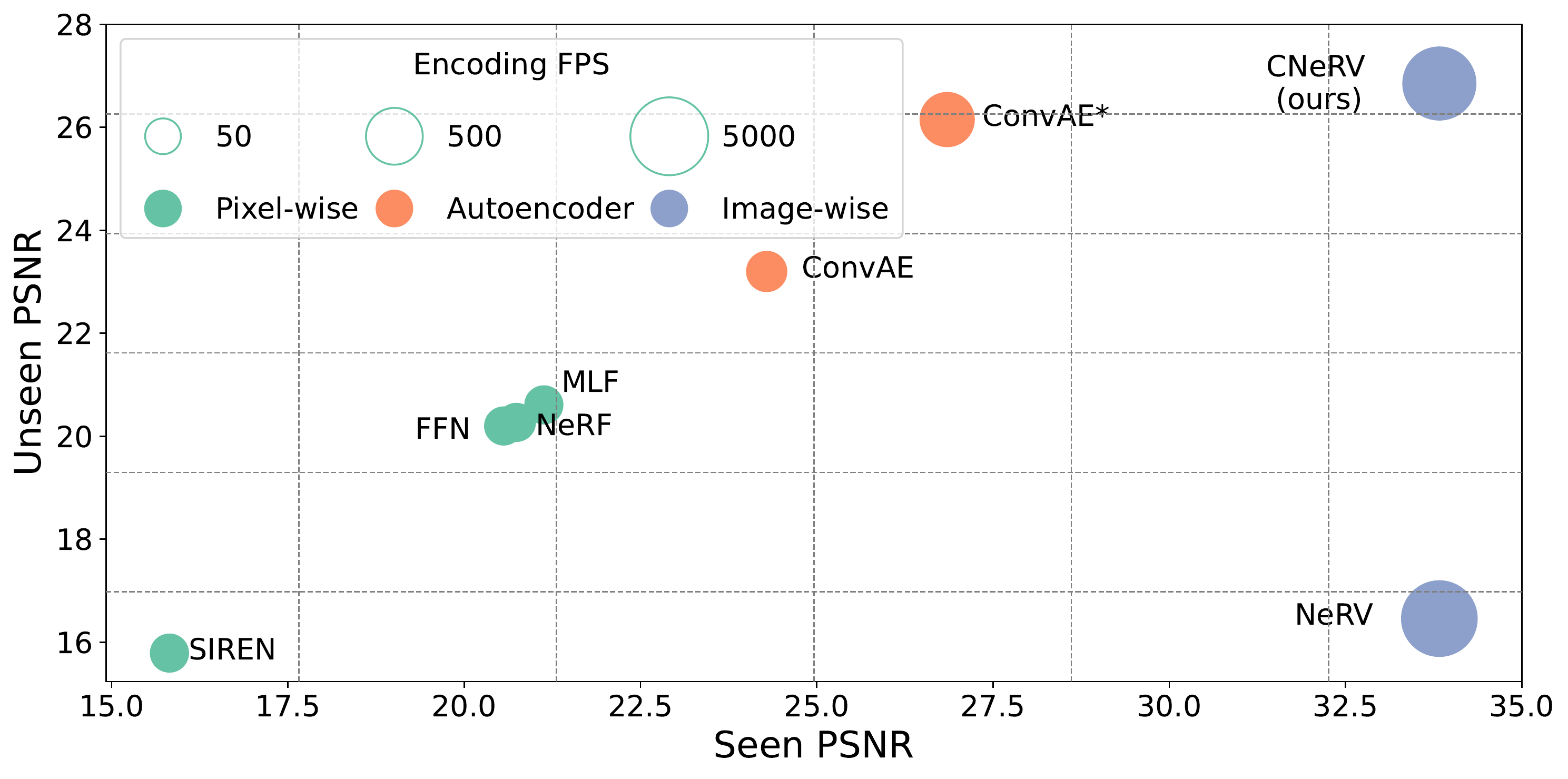}
    \caption{CNeRV achieves the highest unseen PSNR and matches state-of-the-art seen PSNR, at fast encoding speeds. CNeRV and NeRV are image-wise neural representations, ConvAE/ConvAE* are convolutional autoencoders with small/large embedding, respectively; SIREN, FFN, NeRF, and MLF are pixel-wise neural representations.}
    \label{fig:cnerv_teaser}
\end{figure}

Visual data compression remains a fundamental problem in computer vision, and most methods can be seen as autoencoders, consisting of two components: encoder and decoder. 
Traditional compression methods, such as JPEG~\citep{jpeg}, H.264~\citep{h264}, and HEVC~\citep{hevc}, manually design the encoder and decoder based on discrete cosine transform (DCT)~\citep{dct}. 
With the success of deep learning, many attempts~\citep{lu2019dvc,rippel2021elfvc,Agustsson_2020_CVPR,Djelouah_2019_ICCV,Habibian_2019_ICCV,liu2019neural,9247134,Rippel_2019_ICCV,Wu_2018_ECCV} have been made to replace certain components of existing compression pipelines with neural networks. 
Although these learning-based compression methods show high potential in terms of rate-distortion performance, they suffer from  expensive computation, not just to train, but also to encode and decode.
Moreover, as a result of being partially hand-crafted, they are also quite reliant on various hard-coded priors.

To address the heavy computation for autoencoders, implicit neural representations~\citep{pmlr-v97-rahaman19a,sitzmann2020implicit,schwarz2021graf,Chen_2019_CVPR,Park_2019_CVPR} have become popular due to simplicity, compactness, and efficiency. 
These methods show great potential for visual data compression, such as COIN~\citep{dupont2021coin} for image compression, and NeRV~\citep{chen2021nerv} for video compression. 
By representing visual data as neural networks, visual data compression problems can be converted to model compression problems and greatly simplify the complex encoding and decoding pipeline. 

Unlike other implicit methods that map a single network to a single image, NeRV trains as a single network to map timestamps to RGB frames directly for entire videos.
This allows NeRV to achieve incredible results for video compression.
However, because NeRV's input embedding comes from the positional encoding of an image/frame index, which is content-agnostic, NeRV can only \textit{memorize}.
This is evidenced by it achieving surprisingly poor reconstruction quality for unseen data (images/frames that are skipped during training), even when these images only deviate very slightly from images it has seen.
This means that NeRV can only work with a fixed set of images that it has seen during training time, and it could never perform, for example, post-training operations such as frame interpolation.

We thus propose Content-adaptive Neural Representation for Visual Data (CNeRV) to enable internal generalization. 
With a content-adaptive embedding, rather than a temporal/index-based embedding, CNeRV combines the generalizability of autoencoders (AEs) with the simplicity and compactness of implicit representation. 
Similar to implicit representations, CNeRV has a strong decoder, and stores most of the visual prior in the neural network itself. 
Given a tiny embedding, CNeRV can reconstruct the image with high quality, just as NeRV does, and serves as an internally-generalizable neural representation, shown in Figure~\ref{fig:cnerv_teaser}.

We summarize our primary contributions as follows:
\begin{itemize}
    \vspace{-0.6em}
    \item We propose content-adaptive embedding (CAE) to effectively and compactly encode visual information, and generalize to skipped images for a given video/domain.
    \vspace{-0.6em}
    \item  We propose CNeRV based on CAE, which leverages a single-layer mini-network to encode images ($120 \times$ faster than NeRV), with no need for the time-consuming per-image overfitting used by implicit representation methods.
    \vspace{-0.6em}
    \item We demonstrate that CNeRV outperforms autoencoders on the reconstruction task ($+9.5db$ for seen image PSNR, $+3.5db$ for unseen image PSNR).
    \vspace{-0.6em}
    \item We show promising video compression results for both unseen and all frames when compared with traditional visual codec such as H.264 and HEVC.
\end{itemize}

\section{Related Work}

\medskip\noindent\textbf{Neural Representation.}
Implicit neural representations can be divided into two types: pixel-wise representation and image-wise representation. Taking pixel coordinate as input, pixel-wise representation yields outputs based on the input queries, and have become popular for numerous applications, including image reconstruction~\citep{sitzmann2020implicit}, shape regression~\citep{Chen_2019_CVPR,Park_2019_CVPR}, and 3D view synthesis~\citep{schwarz2021graf}. For image-wise implicit representation, NeRV~\citep{chen2021nerv} outputs the whole image given an index, which greatly speeds up the encoding and decoding process compared to pixel-wise methods, and a recent E-NeRV~\citep{li2022nerv} improves the architecture design.
CNeRV is also an image-wise representation method. 
As \citep{mehta2021modulated} points out, most implicit functional representations rely on fitting to each individual test signal, which can be expensive, even with meta-learning~\citep{tancik2021learned} to reduce the amount of regression necessary.
Although we focus entirely on image reconstruction, and its relationship with video-related tasks, we select from these methods some suitable baselines that warrant comparison.
We choose MLP-based methods which leverage (a) periodic activations, such as SIREN~\citep{sitzmann2020implicit} and  MLF (Modulated Local Functional Representations)~\citep{mehta2021modulated}, and (b) Fourier features, such as NeRF~\citep{mildenhall2020nerf} and FFN (Fourier Feature Network)~\citep{tancik2020fourier}.

\medskip\noindent\textbf{Autoencoders.}
Our work is related to other works where a network learns to represent an image in a way that either relies on or later lends itself to reconstruction of the image. 
Of these methods, ours is closely related to auto-encoding~\citep{dct,autoencoder_08,kingma2014autoencoding,NIPS2016_eb86d510}, which focuses on encoding and reconstruction of real images, sometimes by leveraging adversarial techniques~\citep{makhzani2016adversarial,DBLP:journals/corr/DonahueKD16,dumoulin2017adversarially,donahue2019large} that were originally proposed to help synthesize new images~\citep{DBLP:conf/nips/GoodfellowPMXWOCB14}. Numerous algorithmic and architectural improvements~\citep{chen2019log,dupont2018learning,oord2018neural,tolstikhin2019wasserstein} were introduced later based on the vanilla autoencoder. 
We take vanilla convolutional autoencoder and convolutional VAE as baselines in this work.
In fact, our method is akin to an autoencoder that is optimized for data compression: the encoder is very small and fast, while the decoder is reasonably quick and not excessively large.

\medskip\noindent\textbf{Visual Codec.}
Borrowing principles from image compression techniques~\citep{jpeg,jpg2000} and transform coding methods~\citep{dct,antonini:hal-01322224}, traditional video compression methods such as MPEG~\citep{mpeg}, H.264~\citep{h264}, and HEVC~\citep{hevc} are designed to be both fast and accurate.
Recently, deep learning techniques have been proposed to replace portions of the video compression pipeline~\citep{chen2021nerv,rippel2021elfvc,10.1007/978-3-030-58520-4_27,Agustsson_2020_CVPR,Djelouah_2019_ICCV,Habibian_2019_ICCV,liu2019neural,9247134,Rippel_2019_ICCV,Wu_2018_ECCV,khani2021efficient}. 
Although these learning-based methods show promising results for rate-distortion performance, most of them suffer from complex pipelines and heavy computation.
Of all the methods referenced in this section, ours is most closely related NeRV~\citep{chen2021nerv}, which converts frame index to a positional encoding to allow a neural network to memorize and compress a video.
Covered in more detail in Sec.\ \ref{sec:method}, a primary difference from NeRV is CNeRV's use of a content-adaptive embedding, which allows CNeRV to encode frames that were skipped during training time.

\begin{figure*}[t!]
    \centering
    \includegraphics[width=.98\textwidth]{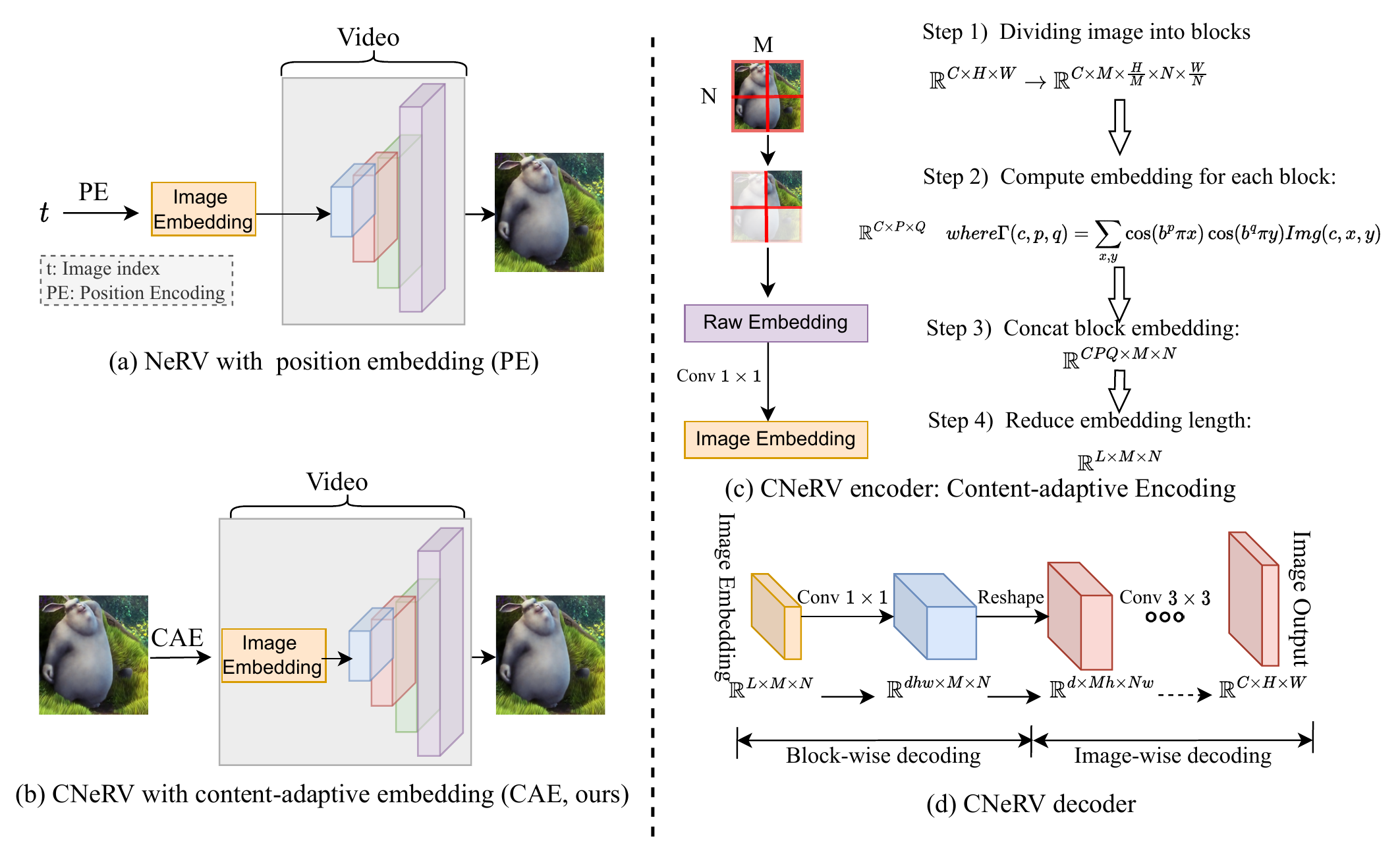}    
    \caption{(a) \textbf{NeRV} represents videos or image datasets as neural neworks, taking an image index as input, and outputting the whole image. (b) \textbf{CNeRV} also stores visual priors in neural networks, but by taking a content-adaptive embedding as input, it can easily generalize internally to unseen (skipped) frames. (c) \textbf{CNeRV encoder} first divides the input image into blocks, then computes content-adaptive embedding, and finally reduce the embedding length with a $1\times1$ convolution. (d) \textbf{CNeRV decoder} consists of block-wise computation ($1\times1$ convolution) and image-wise computation ($3\times3$ convolution), more details are in Figure~\ref{fig:cnerv_archi}. }
    \label{fig:cnerv_idea}
\end{figure*}

\section{Method}\label{sec:method}

Our work on representation and compression for visual data builds on NeRV.
We replace their content-agnostic positional embedding with a proposed content adaptive embedding and a single-layer neural encoder.
To clarify the relationship between CNeRV and NeRV, we first revisit NeRV in Sec.\ \ref{sec:nerv}, then present CNeRV to introduce internal generalization in  Sec.\ \ref{sec:cnerv}, and finally how it can be leveraged for visual data compression in Sec.\ \ref{sec:weight_quant}.

\subsection{Revisiting NeRV}
\label{sec:nerv}
\begin{figure}[t]
    \centering
    \includegraphics[width=.8\linewidth]{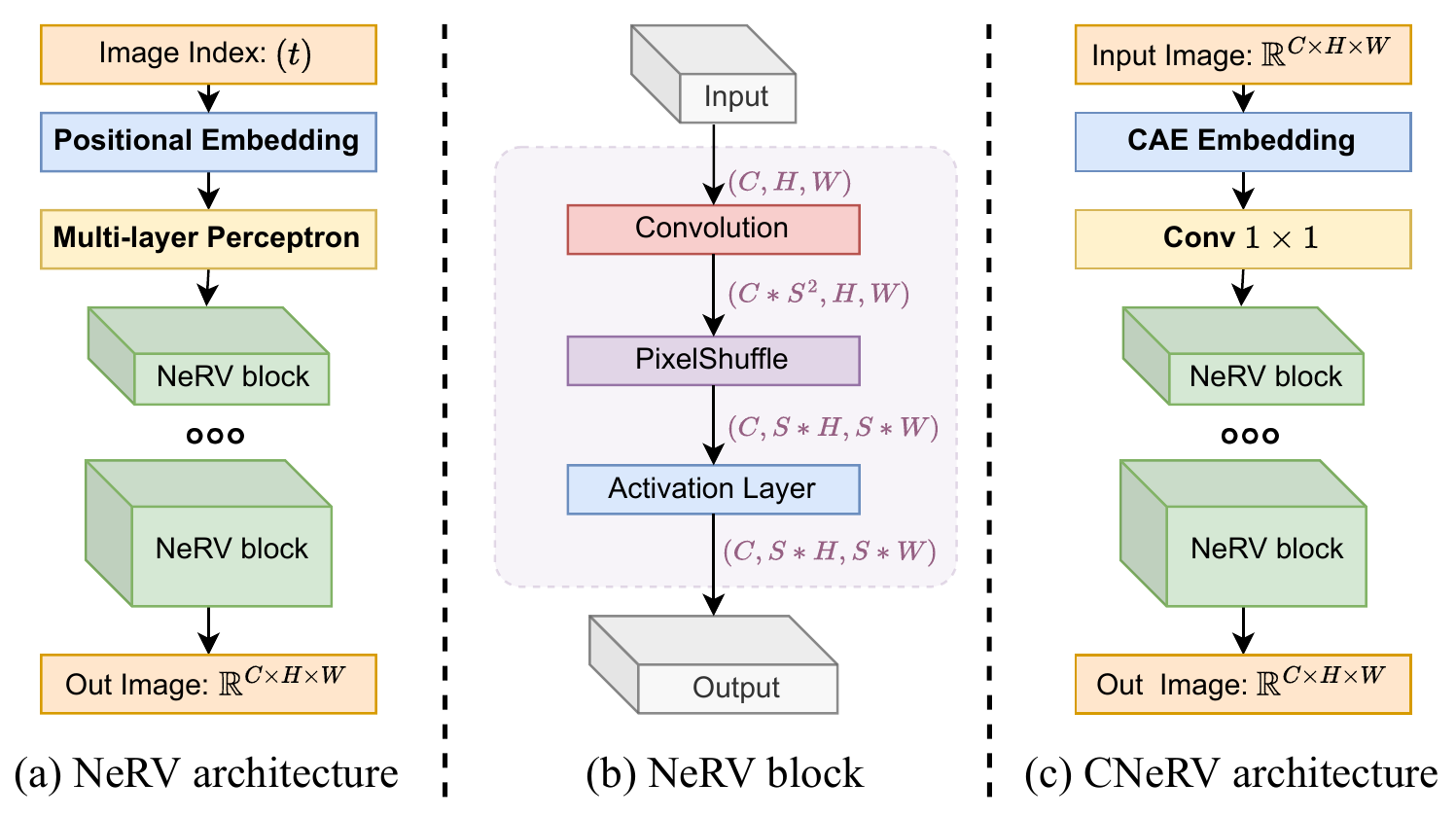}
    \vspace{0.5em}
    \caption{\textbf{Left}: NeRV consists of MLP and NeRV blocks.  \textbf{Middle}: NeRV block. \textbf{Right} CNeRV consists of block-wise computation ($1\times1$ convolution)  and image-wise computation (NeRV blocks with $3\times3$ convolution).}
    \label{fig:cnerv_archi}
    \vspace{-0.5em}
\end{figure}

As shown in Figure~\ref{fig:cnerv_idea}, NeRV takes as input an image index $t$, normalized between $0$ and $1$, and outputs the whole image directly, through an embedding layer and a neural network. The image embedding is given by a positional encoding function:
\begin{equation}
    \footnotesize
    \Gamma(t) = \left(\sin\left(b^0\pi t\right), \cos\left(b^0\pi t\right), \dots, \sin\left(b^{l-1}\pi t\right), \cos\left(b^{l-1}\pi t\right)\right)
    \label{equa:nerv-embed}
\end{equation}
where $t$ is the image index, $b$ is the frequency value, and $l$ is the frequency length.
Specifically, the NeRV network, as illustrated in Figure~\ref{fig:cnerv_archi}(a), consists of a multi-layer perceptron (MLP) and stacked NeRV blocks. To upscale the spatial size, a NeRV block stacks a convolution layer, a pixelshuffle module~\citep{Shi_2016_CVPR}, and an activation layer, as illustrated in Figure~\ref{fig:cnerv_archi}(b).

\subsection{CNeRV: Content-adaptive Neural Representation for Visual Data}
\label{sec:cnerv}

To ameliorate NeRV's lack of internal generalization (results in supplementary material), motivated by its failure to properly correlate neighboring frames, we propose a different embedding.
Rather than a positional encoding, which is content-agnostic, we design a content-aware encoding.
We create a single-layer encoder to perform a learned transform on this encoding and use NeRV decoder for reconstruction.

\noindent\textbf{\underline{CNeRV Encoder}}
We introduce a tiny, single-layer encoder in Figure~\ref{fig:cnerv_idea}(c)  that uses content-adaptive embedding to achieve internal generalization.

\textbf{Dividing image into blocks.}
One insight behind CNeRV is that local visual patterns can be extracted and act as a visual prior. To achieve this, we first divide each input image into blocks and encode them in a batch, illustrated in step $ 1$ at Figure~\ref{fig:cnerv_idea}(c), $\mathbb{R}^{C \times H \times W} \rightarrow \mathbb{R}^{C \times M \times \frac{H}{M} \times N \times \frac{W}{N}}$ where $\mathbb{R}^{C \times H \times W}$ represents the input image, $C$  the image channel, $H$ and $W$ image height and width, $M$ and $N$ block numbers

\textbf{Content-adaptive embedding.}
To improve the embedding generalization, we encode the image  content into an embedding. Inspired by DCT~\citep{dct} and positional encoding, we compute content-adaptive embedding $\mathbb{R}^{C \times P \times Q}$ with
\begin{equation}
    \footnotesize
    \Gamma(c,p,q)=\sum_{x,y} \cos(b^p \pi x)\cos(b^q \pi y)\text{Img}(c,x,y)
    \label{equa:gner-embed}
\end{equation}
where $b$ is the frequency value, $p$ and $q$ are the frequency values, Img($c,x,y$) is the pixel value at location $(x,y)$ for channel $c$, and $x$ and $y$ are normalized between $(0,1)$. 
We then concatenate the block embedding to construct the image embedding $\mathbb{R}^{CPQ \times M\times N}$ as shown in Figure~\ref{fig:cnerv_idea}(c) with steps $2$ and $3$.

\textbf{Reduce embedding dimension.}
Since the raw embedding can be very high in length dimension ($CPQ$), we introduce a $1\times1$ convolution to reduce the length and get an embedding $\mathbb{R}^{L \times M\times N}$. This will be the image latent code and also the input of CNeRV decoder, as illustrated in Figure~\ref{fig:cnerv_idea}(c), step $4$.

\noindent\textbf{\underline{CNeRV Decoder}}
We illustrate the CNeRV decoder in Figure~\ref{fig:cnerv_idea}(d) and divide it into two parts: block-wise and image-wise decoding.

\textbf{Block-wise Decoding.}
Given image embedding $\mathbb{R}^{L \times M\times N}$, we firstly decode it block-wisely with $1\times1$ convolution into $\mathbb{R}^{dhw \times M\times N}$. Then we reshape it into $\mathbb{R}^{d \times Mh\times Nw}$, which means every block embedding $\mathbb{R}^{L \times 1\times 1}$ is decoded into a $\mathbb{R}^{d \times h\times w}$ cube. Since this  is shared by blocks, we refer to it as block-wise decoding.

\textbf{Image-wise Decoding.}
Given $\mathbb{R}^{d \times Mh\times Nw}$, following NeRV~\citep{chen2021nerv}, we upscale the feature map into the final image output $\mathbb{R}^{C \times H\times W}$ with stacked NeRV blocks. Since NeRV block consists of $3\times3$ convolution and fuses information over the whole image, we therefore refer to it as image-wise decoding.

\noindent\textbf{\underline{Loss Objective}}
Following NeRV~\citep{chen2021nerv}, we adopt a combination of L1 and SSIM loss as our loss function for network optimization, following
\begin{equation}
    \footnotesize
    L = \alpha \left\lVert y - v \right\rVert_1 + (1 - \alpha) (1 - \text{SSIM}(y, v))
    \label{equa:loss}
\end{equation}
where $y$ and $v$ are the ground truth image and CNeRV image prediction, and $\alpha$ is a hyperparameter to balance the loss items.

\subsection{Application: Visual Data Compression} \label{sec:weight_quant}
Following NeRV, we use model pruning, quantization, and entropy encoding for model compression. Similar to model quantization, we also apply embedding quantization for visual data compression.
Given a tensor $\mu$, a tensor element $\mu_i$ at position $i$ is
\begin{equation}
    \footnotesize
     \mu_i = \text{Round} ((\mu_i - \mu_\text{min}) / s) * \text{s} + \mu_\text{min}, \text{where} \quad \text{s} = (\mu_\text{max} - \mu_\text{min}) / 2^\text{bit}
     \label{equa:quant}
\end{equation}
`Round' is a function that rounds to the closest integer, `bit' is the bit length for quantization, $\mu_\text{max}$ and $\mu_\text{min}$ are the max and min value for $\mu$, and `scale' is the scaling factor.
Using this equation, each model parameter or frame embedding value can be represented with only `bit' length -- by compressing the model and embedding in this way, we achieve visual data compression. 
When computing quantized model size, we also use entropy encoding as an off-the-shelf technique to save space.

\section{Experiments}\label{sec:experiments}

\noindent\textbf{{\underline{Datasets and implementation details}}}
\label{subsec:implementation-details}
We conduct experiments on both video and image datasets. 
For video, we choose Big Buck Bunny~\citep{bigbuckbunny}~(our default dataset), UVG~\citep{mercat2020uvg},  and MCL-JCL~\citep{wang2016mcl} and list their statistics in Table~\ref{tab:dataset-stat}.
To make it suitable for generation, we crop the video resolution from $1080\times1920$ to $960\times1920$.
To evaluate internal generalization on different resolutions, we downsample video to $480\times960$~(our default resolution) and $240\times480$. 
We hold out 1 in every 5 images/frames for testing, and thus
have a 20\% test split set, referred to as `unseen'.
We also conduct experiments on Celeb-HQ~\citep{CelebAMask-HQ}, a face dataset with 12k/3k images for seen/unseen set, and Oxford Flowers~\citep{nilsback2008automated}, which has 3.3k/1.6k images for seen/unseen set.

For CNeRV architecture on $480 \times 960$ videos, the block number $M \times N$ is $2 \times 4$, frequency value $b$ is $1.15$ and frequency length $P$ and $Q$ are both $15$, the block embedding length $L$ is $60$,  the output size of block-wise decoder $d \times h \times w$ is $620 \times 30 \times60$, followed by 4 NeRV blocks, each with an up-scale factor of 2. By changing channel width $d$, we can build CNeRV with different sizes. For the loss objective, from Equation~\ref{equa:loss}, $\alpha$ is set to $0.7$.  
For NeRV embeddings, we use $b= 1.25$ and $l= 240$ in Equation~\ref{equa:nerv-embed}, following the settings from the original paper. We evaluate the video quality with two metrics: PSNR and MS-SSIM. Bits-per-pixel (BPP) is adopted to evaluate the compression ratio. We implement our method in PyTorch~\citep{NEURIPS2019_9015} and train it in full precision (FP32), on NVIDIA RTX2080Ti. 

We firstly fit the model on the seen split, and evaluate its internal generalization on the unseen set. When computing ``total size'' for a representation method, for implicit representation (\eg NeRV) we only compute model parameters, while for methods with image latent code, we compute both model parameters and image embedding size as total size.
More results can be found in the supplementary material.

\begin{table*}[t]
\centering

\begin{minipage}{.48\textwidth}
    \caption{\textbf{Video dataset} statistics
    }
    \label{tab:dataset-stat}
    \vspace{0.65em}     
    \renewcommand{\tabcolsep}{4pt}
    \resizebox{.99\linewidth}{!}{
    \begin{tabular}{@{}lcccc@{}}
    \toprule
    Dataset & \#frames & \#videos & Duration & FPS \\
    \midrule
    UVG & 3900 & 7 & 5s or 2.5s & 120 \\
    Bunny & 5032 & 1 & $\sim$10min & 8 \\
    MCL & 4115 & 30 & 5s & 24-30 \\
    \bottomrule
    \end{tabular}
    }

    \vspace{1em}

    \caption{Results on different \textbf{video datasets}}
    \label{tab:general-dataset}
    \renewcommand{\tabcolsep}{4pt}    
    \resizebox{.99\linewidth}{!} {
    \begin{tabular}{@{}lcccccc@{}}
    \toprule
    \makecell{Method} & \makecell{Dataset} & \makecell{Embed \\ Length} & \makecell{Total \\ Size} & \makecell{ \\ Seen} & \makecell{PSNR \\ Unseen $\uparrow$} & \makecell{ \\ Gap $\downarrow$} \\
    \midrule
    NeRV & UVG &480 &64M & 36.05 & 23.66 & 12.39 \\
    CNeRV & UVG &480 &64M & 35.83 & \textbf{28.76} & \textbf{7.07}  \\
    \midrule
    NeRV & Bunny &480 &64M & 33.53 & 16.46 & 17.07  \\
    CNeRV & Bunny &480 &64M & 33.83 & \textbf{26.85} & \textbf{6.98}  \\
    \midrule
    NeRV & MCL &480 &64M & 34.83 & 19.44 & 15.39 \\
    CNeRV & MCL &480 &64M & 34.67 & \textbf{26.98} &\textbf{ 7.69}  \\
    \bottomrule
    \end{tabular}
    }
    
    \vspace{1em}
    
    \caption{Results on different \textbf{model sizes} }
    \label{tab:general-model-size}
    \resizebox{.99\linewidth}{!} {
    \begin{tabular}{@{}lcccccc@{}}
    \toprule
    \makecell{Method} & \makecell{Model \\ Size} & \makecell{Embed \\ Length} & \makecell{Total \\ Size} & \makecell{ \\ Seen} & \makecell{PSNR \\ Unseen $\uparrow$} & \makecell{ \\ Gap $\downarrow$} \\
    \midrule
    NeRV & Small &480 &32M & 31 & 16.72 & 14.28 \\
    CNeRV & Small &480 &32M & 31.33 & \textbf{26.41} & \textbf{4.92}  \\
    \midrule
    NeRV & Medium &480 &64M & 33.53 & 16.46 & 17.07 \\
    CNeRV & Medium &480 &64M & 33.83 & \textbf{26.85} & \textbf{6.98}  \\
    \midrule
    NeRV & Large &480 &97M & 35.32 & 16.04 & 19.28  \\
    CNeRV & Large &480 &97M & 35.5 & \textbf{27.08} & \textbf{8.42}  \\
    \bottomrule
    \end{tabular}
    }

\end{minipage}
\hfill
\begin{minipage}{.48\textwidth}
    \renewcommand{\tabcolsep}{4pt}    
    \caption{Results on \textbf{video resolutions}}
    \label{tab:general-video-resolution}
    \vspace{0.65em}
    \resizebox{.99\linewidth}{!} {
    \begin{tabular}{@{}lcccccc@{}}
    \toprule
    \makecell{Method} & \makecell{Video \\ Resolution} & \makecell{Embed \\ Length} & \makecell{Total \\ Size} & \makecell{ \\ Seen} & \makecell{PSNR \\ Unseen $\uparrow$} & \makecell{ \\ Gap $\downarrow$} \\
    \midrule
    NeRV & 240*480 & 480 & 60M & 37.14 & 16.9 & 20.24  \\
    CNeRV & 240*480 & 480 & 60M & 36.99 & \textbf{27.97} & \textbf{9.02} \\
    \midrule
    NeRV & 480*960 & 480 & 64M & 33.53 & 16.46 & 17.07  \\
    CNeRV & 480*960 & 480 & 64M & 33.83 & \textbf{26.85} & \textbf{6.98}  \\
    \midrule
    NeRV & 960*1920 & 480 & 67M & 32.06 & 16.06 & 16  \\
    CNeRV & 960*1920 & 480 & 67M & 32.4 & \textbf{26.15} & \textbf{6.25}  \\
    \bottomrule
    \end{tabular}
    }

    \vspace{0.6em}

    \caption{Results on \textbf{training data size}}
    \label{tab:general-data-size}
    \resizebox{.99\linewidth}{!} {
    \begin{tabular}{@{}lcccccc@{}}
    \toprule
    \makecell{Method} & \makecell{Training \\ Images} & \makecell{Embed \\ Length} & \makecell{Total \\ Size} & \makecell{ \\ Seen} & \makecell{PSNR \\ Unseen $\uparrow$} & \makecell{ \\ Gap $\downarrow$}  \\
    \midrule
    NeRV & ~1k & 480 & 64M & 34.6 & 12.57 & 22.03  \\
    CNeRV & ~1k & 480 & 64M & 34.78 & \textbf{26.41} & \textbf{8.37}  \\
    \midrule
    NeRV & ~2k & 480 & 64M & 33.53 & 16.46 & 17.07  \\
    CNeRV & ~2k & 480 & 64M & 33.83 &\textbf{ 26.85} & \textbf{6.98} \\
    \midrule
    NeRV & ~4k & 480 & 64M & 32.78 & 20.68 & 12.1 \\
    CNeRV & ~4k & 480 & 64M & 32.94 & \textbf{27.75} & \textbf{5.19} \\
    \bottomrule
    \end{tabular}
    }
    
    \vspace{0.6em}
    
    \caption{Results on \textbf{image datasets}}
    \label{tab:general-image-dataset}
    \renewcommand{\tabcolsep}{4pt}
    \resizebox{.99\linewidth}{!} {
    \begin{tabular}{@{}lcccccc@{}}
    \toprule
    \makecell{Method} & \makecell{Dataset} & \makecell{Embed \\ Length} & \makecell{Total \\ Size} & \makecell{ \\ Seen} & \makecell{PSNR \\ Unseen $\uparrow$} & \makecell{ \\ Gap $\downarrow$} \\
    \midrule
    NeRV & Celeb &240 &33M & 27.44 & 11.27 & 16.17  \\
    CNeRV & Celeb &240 &33M & 27.42 & \textbf{21.34} & \textbf{6.08} \\
    \midrule
    NeRV & Flower &240 &35M & 27 & 11.29 & 15.71  \\
    CNeRV & Flower &240 &36M & 27.04 & \textbf{18.54} & \textbf{8.5} \\
    \bottomrule
    \end{tabular}
    }
\end{minipage}
\end{table*}

\noindent\textbf{{\underline{Comparison to NeRV}}}
We present our main result in Table~\ref{tab:general-dataset}, which shows that CNeRV consistently outperforms NeRV in terms of both reconstruction quality of unseen images (unseen PSNR) and internal generalizability (PSNR gap between seen and unseen frames/images). Extended tables with more details and results in this section can be found in the supplementary material.
Note that the generalization of NeRV becomes worse when FPS decreases, \ie with diversified frame content, while CNeRV retains its generalizability in all cases.
We first verify CNeRV's superior performance holds under a variety of settings, increasing model size in Table~\ref{tab:general-model-size}, various video resolution in Table~\ref{tab:general-video-resolution}, various training images in Table~\ref{tab:general-data-size}.

We then extend our finding on the non-sequential nature of NeRV's representation to apply it to an image dataset, using an arbitrary index for each image as the frame index.
NeRV thus takes image index (1/N, ..., N/N respectively for N images) as input.
With this adaptation, we compare them in Table~\ref{tab:general-image-dataset} and CNeRV shows superior internal generalization.

There is no question that NeRV and CNeRV both succeed by fitting to their training data.
However, as Figure~\ref{fig:cnerv_vs_nerv} shows, CNeRV is able to do this without sacrificing generalization for increasing training epochs (it does not exhibit the same overfitting behavior).
CNeRV's performance for unseen images doesn't decrease as training time increases, even though its seen image reconstruction quality continues to improve.
Furthermore, Table~\ref{tab:encode-time} points out that CNeRV is vastly superior in terms of \textbf{encoding time} for unseen frames, when accounting for the fact that NeRV must be fine-tuned on those previously unseen images to reach competitive PSNR. 
Thus, CNeRV's internal generalization allows it to save training/encoding time.

\begin{table}[t!]
\centering
\caption{Comparison of \textbf{encoding time}. We fine-tune NeRV on unseen images until it reaches comparable PSNR with CNeRV.}
\label{tab:encode-time}
    \vspace{0.65em}
    \renewcommand{\tabcolsep}{6pt}
    \resizebox{.8\linewidth}{!} {
    \begin{tabular}{@{}lccccccc@{}}
    \toprule
     \multirow{2}{*}{Method} &  \multirow{2}{*}{Overfit} & \multicolumn{2}{c}{UVG} & \multicolumn{2}{c}{Bunny} & \multicolumn{2}{c}{MCL} \\
      \cmidrule(lr){3-4} \cmidrule(lr){5-6} \cmidrule(lr){7-8} 
     & & PSNR $\uparrow$  & FPS $\uparrow$ & PSNR $\uparrow$   & FPS $\uparrow$ & PSNR $\uparrow$   & FPS $\uparrow$ \\
     \midrule
    NeRV & \checkmark & 28.69  &0.13 & 26.59  &0.14 & 26.75  & 0.13 \\
    CNeRV & & 28.76  & \textbf{16.1} ($124\times$) & 26.85  &\textbf{16.1} ($115\times$) & 26.98 &\textbf{16.1} ($124\times$) \\
    \bottomrule
    \end{tabular}
    }
\end{table}

\begin{table}[t!]
\centering
    
\caption{Compare with \textbf{Autoencoders} and \textbf{pixel-wise neural representations} 
} 
\label{tab:autoencoder}
\vspace{0.65em}
\resizebox{.7\linewidth}{!} {
    \renewcommand{\tabcolsep}{4pt}
    \begin{tabular}{@{}l|cccccccc@{}}
    \toprule
    Methods & \makecell{Image-\\wise} & \makecell{Embed \\ Length} & \makecell{Total \\ Size} & \makecell{Training \\ time} & \makecell{PSNR\\ seen $\uparrow$} & \makecell{PSNR\\ unseen $\uparrow$} & \makecell{Encoder \\ size $\downarrow$} & \makecell{Encoding \\ time $\downarrow$}\\
    \toprule
    SIREN~\citep{sitzmann2020implicit} &  & 480 & 66M& $2\times$ &15.82 & 15.79 & 62M & 16.9ms \\
    FFN~\citep{tancik2020fourier} &  & 480 & 66M&$2\times$ &20.69 & 20.33 & 62M & 16.9ms \\
    NeRF~\citep{mildenhall2020nerf} &  & 480 &66M& $2\times$ & 21.04 & 20.56 & 62M & 16.9ms \\
    MLF~\citep{mehta2021modulated} &  & 480  &66M& $2\times$ & 21.13 & 20.61 & 62M & 16.9ms \\
    \midrule
    ConvAE & \checkmark & 480 &68M& $2\times$ &24.29 & 23.2 & 47M & 13.4ms \\
    ConvVAE & \checkmark & 480 &68M& $2\times$ &23.92 & 21.71 & 46M & 13.7ms \\
    ConvAE* & \checkmark & \textbf{12k} &68M& $2\times$ &26.83 & 26.15 & 1.9M & 2.9ms \\
    \arrayrulecolor{lightgray}
    \cmidrule{1-9}
    CNeRV (ours) & \checkmark & 480 &64M& $1/6\times$ & 24.35 & 23.19 & 0.4M & 0.37ms \\
    CNeRV (ours) & \checkmark & 480 & 64M & $1\times$ & \textbf{33.83} & \textbf{26.85} & \textbf{0.4M} &
    \textbf{0.37ms} \\
    \arrayrulecolor{black}
    \bottomrule
    \end{tabular}
}
\end{table}

\begin{table}[t!]
    \centering
\begin{minipage}[]{.59\textwidth}
\centering
     \includegraphics[width=.98\linewidth]{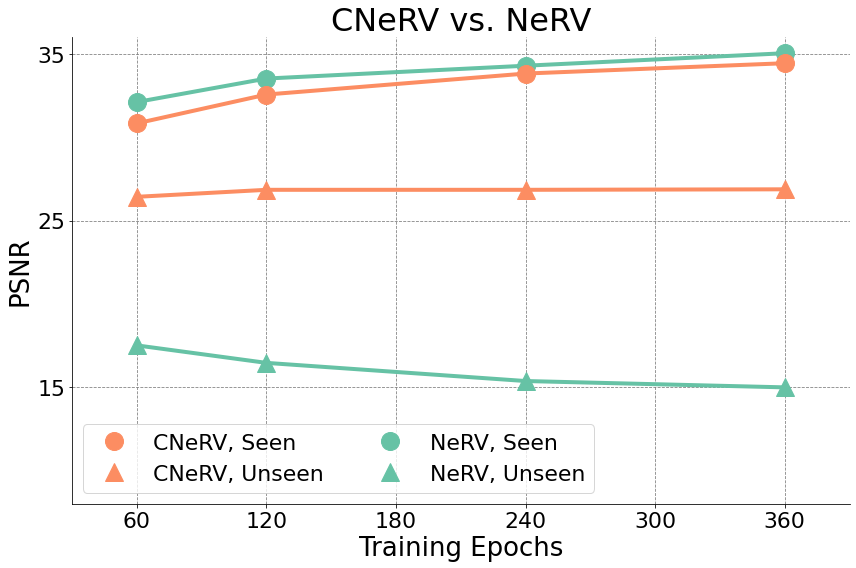}
    \vspace{0.5em}
    \captionof{figure}{CNeRV keeps increasing seen PSNR and remains stable on unseen PSNR while NeRV's better seen PSNR comes at the cost of unseen PSNR
    }
    \label{fig:cnerv_vs_nerv}
\end{minipage}    
\hfill
\begin{minipage}[]{.4\textwidth}
\centering
    \caption{Decoding speed, frames per second (FPS) reported}
    \label{tab:decode-speed}    
    \vspace{0.5em}
        \footnotesize
    \begin{tabular}{@{}c|ccc@{}}
        \toprule
         Resolution & H.264 & H.265 & CNeRV  \\
         \midrule
         $480\times960$ & 85.4 & 56.3 & \textbf{123.4} \\
         $960\times1920$ & 24.5 & 16.2 & \textbf{31.6} \\
         \bottomrule
    \end{tabular}
    
\vspace{1em}
    
\caption{\textbf{Interpolation} results on different datasets. We show PSNR of unseen images with interpolated and ground truth embedding}
\label{tab:main-interpolate}
\vspace{0.5em}
\footnotesize
    \renewcommand{\tabcolsep}{2pt}
\begin{tabular}{@{}lcc@{}}
\toprule

Dataset & \makecell{GT \\ embedding} & \makecell{Embedding \\ interpolation } \\
\midrule
UVG & 28.76 & 28.88 \\
MCL & 26.85 & 26.33 \\
Bunny & 26.98 & 24.94 \\
\bottomrule
\end{tabular}

\end{minipage}

\end{table}

\noindent\textbf{{\underline{Comparison with Other Reconstruction Methods}}}
We also compare with autoencoders and pixel-wise neural representations. 
For autoencoders, we choose the most common convolutional autoencoder and convolutional variational autoencoder as baselines, referred to as ConvAE and ConvVAE. 
They reduce the image into the same block number as CNeRV (\ie $2\times4$) with strided convolution. 
The block embedding length is the same with CNeRV (\ie $60$) as well. 
We also compare with ConvAE* which light neural networks where most visual information is stored in the huge image-specific embedding.  
For pixel-wise neural representations, we choose  NeRF~\citep{mildenhall2020nerf}, SIREN~\citep{sitzmann2020implicit}, FFN~\citep{tancik2020fourier}, and MLF~\citep{mehta2021modulated} as baselines. 
Following MLF~\citep{mehta2021modulated}, we train a separate auto-encoder to provide content information besides the coordinate input.
For fair comparison, we keep the latent code the same length as CNeRV.
All other setups also follow MLF~\citep{mehta2021modulated}.

Table~\ref{tab:autoencoder} shows reconstruction results for both seen and unseen images, as well as encoder model size and encoding speed. 
Note that for encoding time, we only consider the forward time, ignoring the data loading overhead.
With a tiny encoder and strong decoder, CNeRV outperforms autoencoders and pixel-wise neural representations in many regards, including reconstruction quality of seen and unseen images, and encoding speed.
We show visualization results for seen images in Figure~\ref{fig:seen-visulization} and unseen images in Figure~\ref{fig:unseen-visulization}.
With the same embedding length and similar total size, CNeRV outperforms NeRV with better detail, absence of artifacts, and lack of spillover from previous frames.
Although autoencoder with large embeddings can reach comparable generalization for unseen images, it struggles a lot for reconstruction of seen images.
But, visual differences still exist between CNeRV and ground truth, and future work can focus on mitigating these issues.

\begin{figure*}[t!]
    \centering
    \includegraphics[width=.95\linewidth]{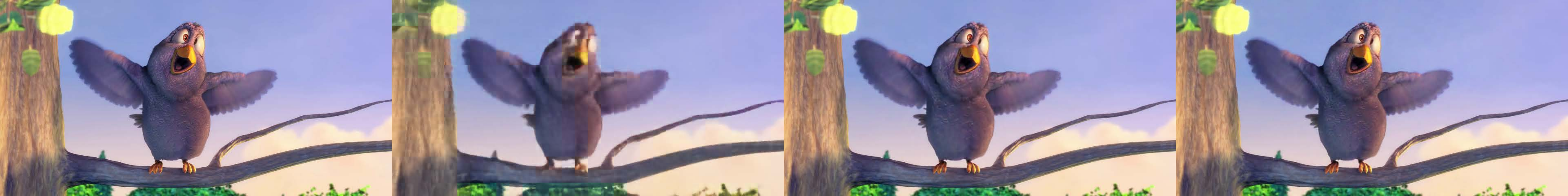}
    \includegraphics[width=.98\linewidth]{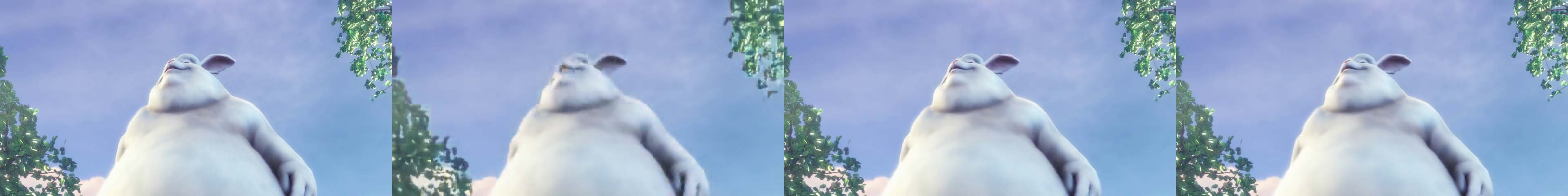}
    \includegraphics[width=.98\linewidth]{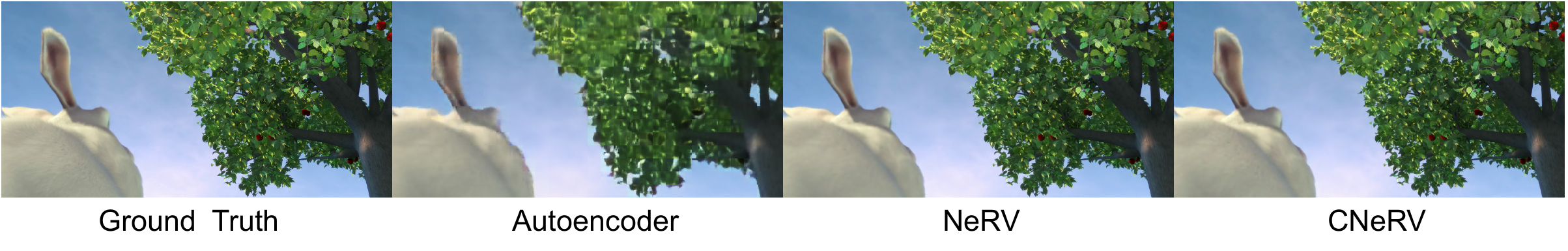}
    \vspace{0.3em}
    \caption{Video reconstruction for seen frames from video Bunny. CNeRV results are comparable to NeRV, while autoencoders suffers from bluriness, due to the fact that without implicit representation it cannot perform well in the compression setting (tiny embedding).}
    \label{fig:seen-visulization}
    \vspace{1em}

    \includegraphics[width=.95\linewidth]{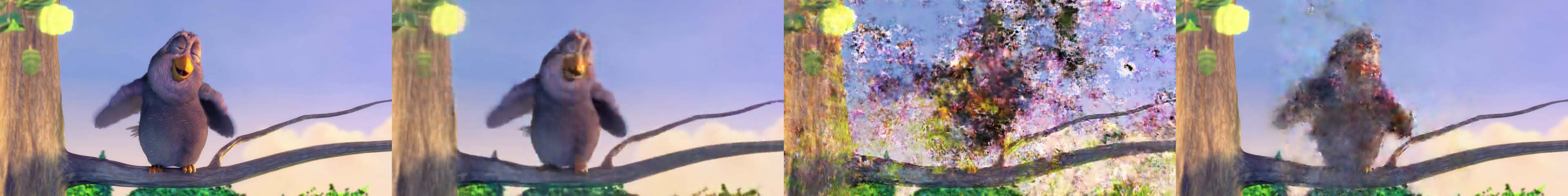}
    \includegraphics[width=.98\linewidth]{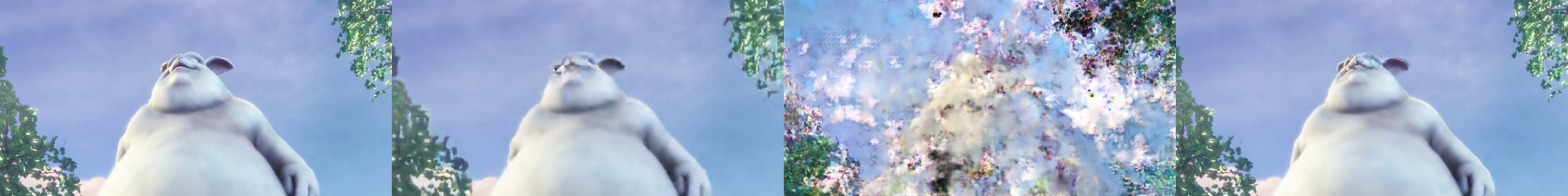}
    \includegraphics[width=.98\linewidth]{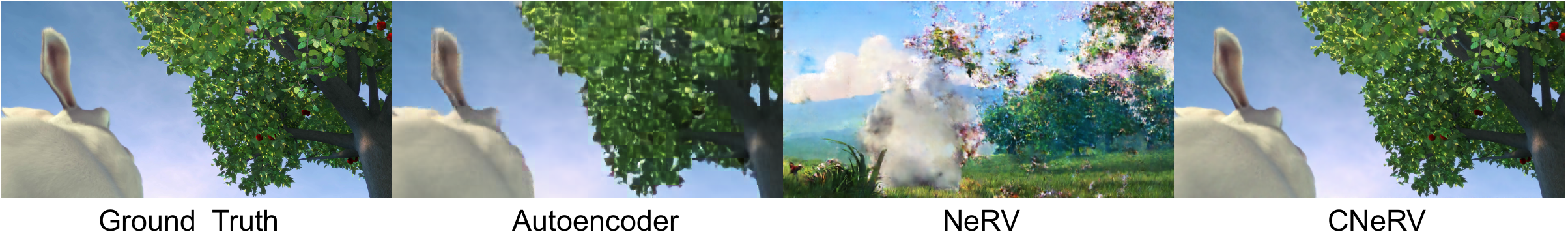}
    \vspace{0.3em}
    \caption{Video reconstruction for unseen frames from Bunny. CNeRV results are similar for unseen as for seen, while autoencoders is still significantly blurrier than CNeRV. NeRV suffers from a myriad of failures.}
    \label{fig:unseen-visulization}
    \vspace{-1em}
\end{figure*}

For low reconstruction quality of pixel-wise neural representations, we believe both these methods are designed and optimized for low-resolution images, and a similar performance drop on high-resolution images is also observed in MLF~\citep{mehta2021modulated}.
For the ConvAE and ConvVAE, we speculate two potential causes for the inferior reconstruction capacity.
First, due to the fact that the parameters are more evenly balanced between encoder and decoder, they rely on larger embeddings and struggle for our embedding size.
Second, we speculate that small training datasets likely limit their capability as well.
We show in the supplementary material that increasing model size, embedding length, and amount of training data each make autoencoders more competitive with CNeRV.

We also compare decoding speed with traditional codecs in Table~\ref{tab:decode-speed}. 
The decoding of traditional codecs are measured with 8 CPUs, while CNeRV is measured on 1 RTX2080ti.
As a video neural representation, CNeRV shows good decoding advantage due to its simplicity and can be deployed easily.

\begin{figure}[t!]
\begin{minipage} {.49\linewidth}
    \includegraphics[width=.98\linewidth]{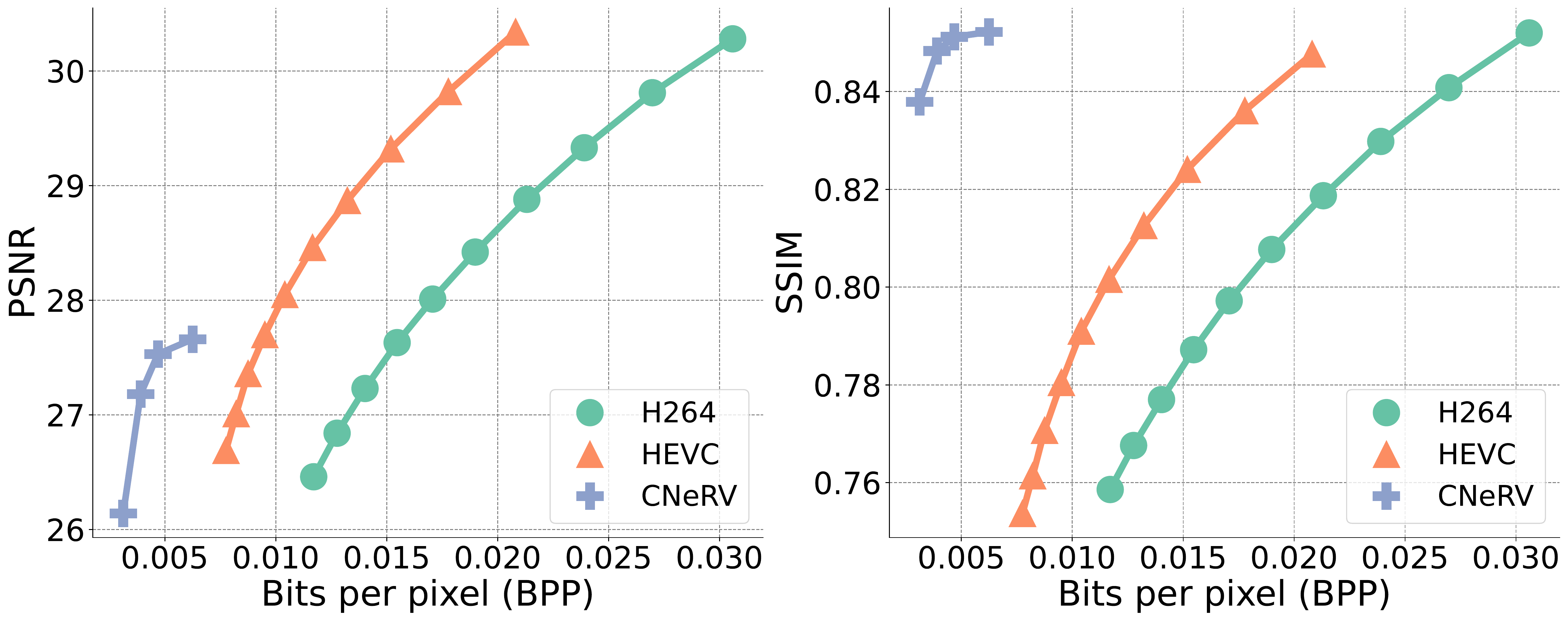}
    \vspace{0.65em}
    \caption{Compression for \textbf{unseen} frames.
    }
    \label{fig:rate_distortion_unseen}
\end{minipage}    
\hfill
\vspace{1em}
\begin{minipage} {.49\linewidth}
    \includegraphics[width=.98\linewidth]{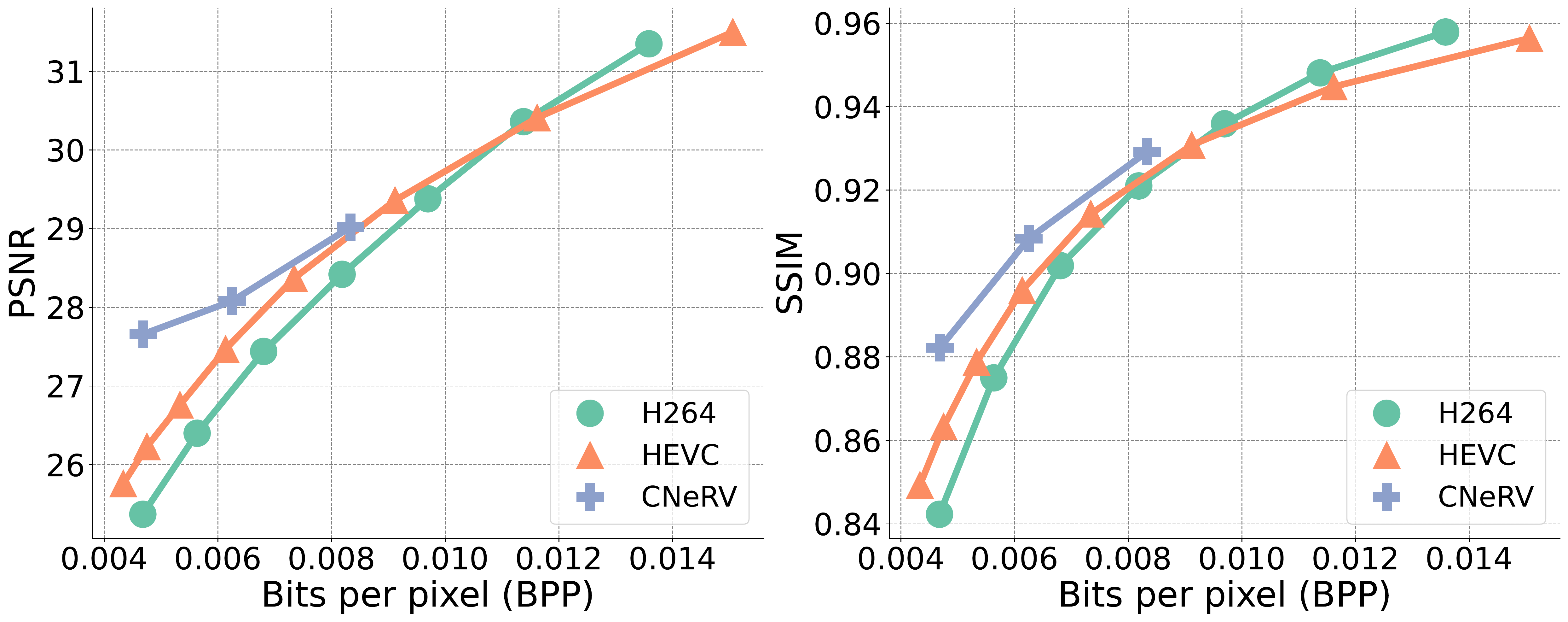}
    \vspace{0.65em}
    \caption{Compression for \textbf{all} frames.
    }
    \label{fig:rate_distortion_all}
\end{minipage}   
\vspace{-2em}
\end{figure}

\noindent\textbf{{\underline{Frame interpolation and Visualization}}}
Given neighboring frames are typically similar, we investigate whether using CNeRV to encode unseen frames is better than interpolating from the embeddings of the neighboring seen frames.
We show these results in Table~\ref{tab:main-interpolate}.
Interpolated embeddings achieve similar performance with actual CNeRV embeddings for unseen images across the lower FPS datasets, but significantly lower results for the Bunny dataset.
With the same embedding length and similar total size, CNeRV outperforms NeRV and autoencoder with better detail for both seen and unseen frames.

\noindent\textbf{{\underline{Visual Data Compression}}}
We also show visual data compression results for CNeRV, as discussed in Sec.\ \ref{sec:weight_quant}.
We compare the rate-distortion results with traditional visual codec such as H.264 and HEVC.
For unseen frames, we compare visual comparison results in Figure~\ref{fig:rate_distortion_unseen} where CNeRV bitrates only consider image embedding as all other autoencoder methods do~\citep{rippel2021elfvc,10.1007/978-3-030-58520-4_27,Agustsson_2020_CVPR,Djelouah_2019_ICCV}.  
Besides, we evaluate compression results on the full video (both seen and unseen frames) frames where we combine both the image embedding and model parameters to compute bitrates, CNeRV outperforms H.264 and HEVC on both PSNR and SSIM with similar bpp in Figure~\ref{fig:rate_distortion_all}.

\section{Conclusion}

In this work, we propose a content-adaptive neural representation, CNeRV. 
CNeRV combines the generalizability of autoencoders and simplicity and compactness of neural representation.
With a single-layer mini-encoder to generate the embedding, CNeRV outperforms autoencoders for the reconstruction task in terms of image quality ($+3.5db$ for unseen image PSNR), encoding speed ($ 36\times$ faster), and encoder size ($ 116\times$ smaller).
We leverage this content-adaptive embedding with CNeRV to encode unseen images quickly ($120 \times$ faster than NeRV), with no need for the time-consuming per-image overfitting.
We also show  promising visual data compression results and provide embedding analysis.

\textbf{Acknowledgement.} This project was partially funded by an independent grant from Facebook AI and the DARPA SAIL-ON (W911NF2020009) program.

\bibliography{egbib}

\appendix

\section{More Results}
\subsection{Embedding Quantization Results}
\begin{figure}[h!]
    \centering
     \includegraphics[width=.6\linewidth]{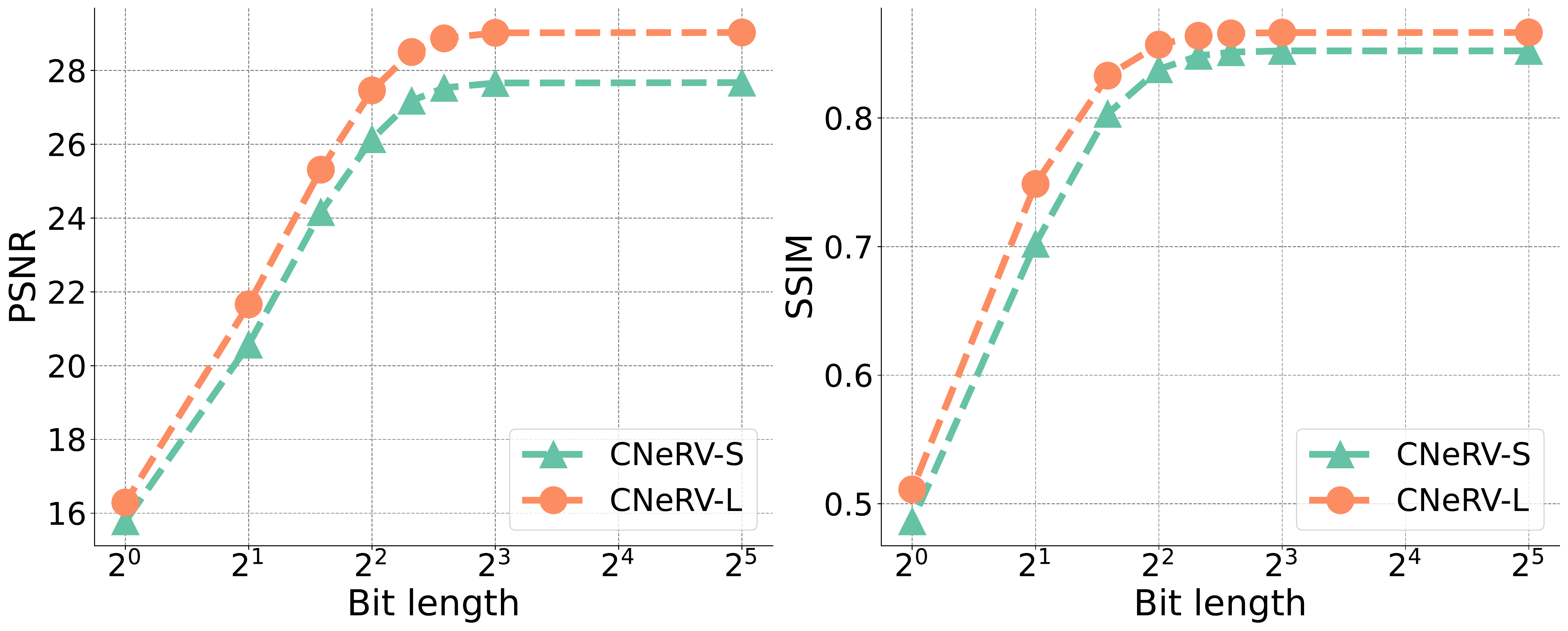}
    \vspace{0.65em}
    \caption{\textbf{Embedding quantization}, We evaluate reconstruction quality on unseen images with embedding quantization.}
    \label{fig:embed-quant}
\end{figure}
For unseen frames and with an $8$-bit quantized model, we further quantize the block embedding, ranging from the original $32$ bit to $1$ bit, where where the image embedding can maintain most of its capacity with only $6$ bits in Figure~\ref{fig:embed-quant}

\subsection{More Autoencoder Results}
We first provide more autoencoder results, where both seen PSNR and unseen PSNR improve as we increase the training images, embedding length, and model size. 

\begin{table}[h]
\centering
\caption{Autoencoder results with more training images, bigger model size, and longer embeddings.}
\begin{tabular}{@{}l|ccccc@{}}
\toprule
Methods  & \makecell{Embed \\ Length} & \makecell{Model \\ Size} & \makecell{Training \\ Images} & \makecell{PSNR\\ Seen $\uparrow$} & \makecell{PSNR\\ Unseen $\uparrow$} \\
\toprule
ConvAE & 480 & 68M & 2k & 24.29 & 23.2 \\
ConvAE & 1000 & 68M & 4k & 26.6 & 25.92 \\
ConvAE & 48k & 83M & 4k & \textbf{29.28} & \textbf{29.28} \\
\bottomrule
\end{tabular}
\end{table}

\subsection{Sensitivity Analysis and Limitations}
\label{subsec:exp-ablations}

\begin{table*}[h!]
    \renewcommand{\tabcolsep}{4pt}
    \begin{minipage}[t]{0.49\textwidth}
        \centering
        \caption{\textbf{Frequency value $b$} ablation }
        \label{tab:cae-base-ablation}
        \vspace{0.5em}
        \resizebox{.98\linewidth}{!} {
        \begin{tabular}{@{}ccccc@{}}
        \toprule
        \multirow{2}{*}{\makecell{Base \\ number}} & \multicolumn{2}{c}{PSNR} & \multicolumn{2}{c}{MS-SSIM} \\
        \cmidrule(lr){2-3} \cmidrule(l){4-5} 
          & Seen $\uparrow$ & Unseen $\uparrow$  & Seen $\uparrow$ & Unseen $\uparrow$ \\
          \midrule
        1.05& 32.6 & 25.94  & 0.915 & 0.8072  \\
        1.15& \textbf{33.83} & \textbf{26.85}  & \textbf{0.9539} &\textbf{ 0.8242} \\
        1.25& 33.5 & 26.67 & 0.949 & 0.8217\\
        \bottomrule
        \end{tabular}
        }
    \end{minipage}
    \hfill
    \begin{minipage}[t]{0.49\textwidth}
     \caption{\textbf{Frequency length} ablation}
    \label{tab:cae-freq-num-ablation}    
    \vspace{0.5em}
        \resizebox{.98\linewidth}{!} {
        \begin{tabular}{@{}ccccc@{}}
        \toprule
        \multirow{2}{*}{\makecell{Freq \\ number}} & \multicolumn{2}{c}{PSNR} & \multicolumn{2}{c}{MS-SSIM} \\
        \cmidrule(lr){2-3} \cmidrule(l){4-5} 
          & Seen $\uparrow$ & Unseen $\uparrow$ & Seen $\uparrow$ & Unseen $\uparrow$  \\
          \midrule
        10& 32.75 & 26.3  & 0.9229 & 0.8151  \\
        15& 33.83 & \textbf{26.85}  & 0.9539 & \textbf{0.8242}  \\
        20& \textbf{34.07} & 26.84  & \textbf{0.9565} & 0.8242  \\
         \bottomrule
         \end{tabular}
     }
    \end{minipage}
\end{table*}

\begin{figure}
    \centering
     \caption{Sensitivitiy for \textbf{Block number}}
    \label{tab:cae-block-num-ablation}
    \vspace{0.5em}
        \resizebox{.6\linewidth}{!} {
        \begin{tabular}{@{}cccccc@{}}
        \toprule
        \multirow{2}{*}{\makecell{Block \\ number}} & \multirow{2}{*}{\makecell{Embed \\ length}} & \multicolumn{2}{c}{PSNR} & \multicolumn{2}{c}{MS-SSIM} \\
                \cmidrule(lr){3-4} \cmidrule(l){5-6} 
        &  & Seen $\uparrow$ & Unseen $\uparrow$ &  Seen $\uparrow$ & Unseen $\uparrow$ \\
          \midrule
        $1\times2$ & 480 & 30.22 & 25.31  & 0.8975 & 0.8059  \\
        $2\times4$ & 480 &\textbf{ 33.83 }& \textbf{26.85}  & \textbf{0.9539} & \textbf{0.8242}  \\
        $5\times10$ & 500 & 4.7 & 4.69  & 0.3344 &0.3336 \\
        \midrule
        $5\times10$ & 1000 & 34.49 & 27.63   &  0.9788 & 0.8597 \\
        \bottomrule
        \end{tabular}
        }
\end{figure}

We conduct an ablation study on our content-adaptive encoding.
For base value $b$ in Equation~\ref{equa:gner-embed}, Table~\ref{tab:cae-base-ablation} shows that $1.15$ performs better than $1.05$ and $1.25$. 
For frequency length $P$ and $Q$,  Table~\ref{tab:cae-freq-num-ablation} shows results with $10$, $15$, and $20$. 
Although  $20$ is better for seen PSNR, it does not further improve unseen PSNR and will introduce more encoding computation.
Since we mainly focus on internal generalizability in this work, \ie unseen image PSNR, we choose $15$ as the default frequency length.
For block numbers, Table~\ref{tab:cae-block-num-ablation} shows results for $1\times2$, $2\times4$, and $5\times10$. 
where total embedding length is computed by $M \times N \times L$. 
With $2\times4$ blocks, CNeRV reaches the best performance for both seen and unseen images. 
Note that when embedding length is too short (\eg 10 for block number $5\times10$), CNeRV fails to overfit; nevertheless, it can still perform reasonable reconstruction.

\noindent\textbf{Limitation}
The main limitation of our method is that with a small training dataset, unseen PSNR still lags behind seen PSNR. 
Although more training images can alleviate this problem, CNeRV still cannot reconstruct unseen images with perfect fidelity in this work. 
This also limits its application for visual codec or compression methods.

\subsection{More visualization results}
\begin{figure*}
    \centering
    \includegraphics[width=.98\linewidth]{figures/vis/seen_pred_79_gt_ae_nerv_gnerv_28.24_37.11_37.23.pdf}
    \includegraphics[width=.98\linewidth]{figures/vis/seen_pred_206_gt_ae_nerv_gnerv_28.61_40.28_40.3.pdf}
    \includegraphics[width=.98\linewidth]{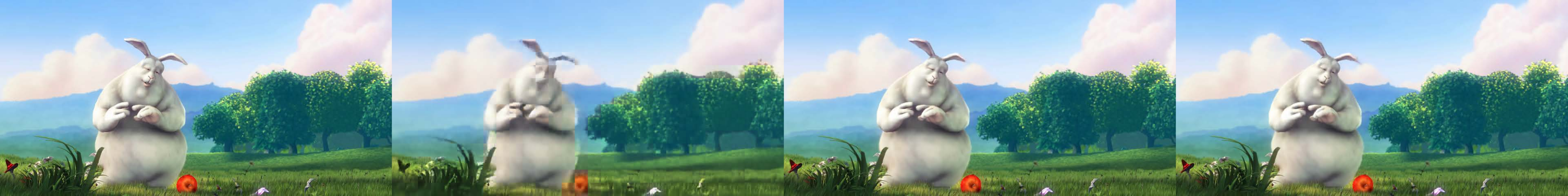}
    \includegraphics[width=.98\linewidth]{figures/vis/seen_pred_302_gt_ae_nerv_gnerv_23.63_32.5_32.2_full.pdf}

    \caption{Video compression results for seen frames from video Bunny. Note that CNeRV results are comparable to NeRV, while autoencoders suffers from bluriness, due to the fact that without implicit representation it cannot perform well in the compression setting (tiny embedding).}
    \label{fig:seen-visulization}
\end{figure*}

\begin{figure*}
    \centering
    \includegraphics[width=.98\linewidth]{figures/vis/unseen_pred_38_gt_ae_nerv_gnerv_28.69_14.11_24.88.pdf}
    \includegraphics[width=.98\linewidth]{figures/vis/unseen_pred_103_gt_ae_nerv_gnerv_28.43_15.08_35.3.pdf}
    \includegraphics[width=.98\linewidth]{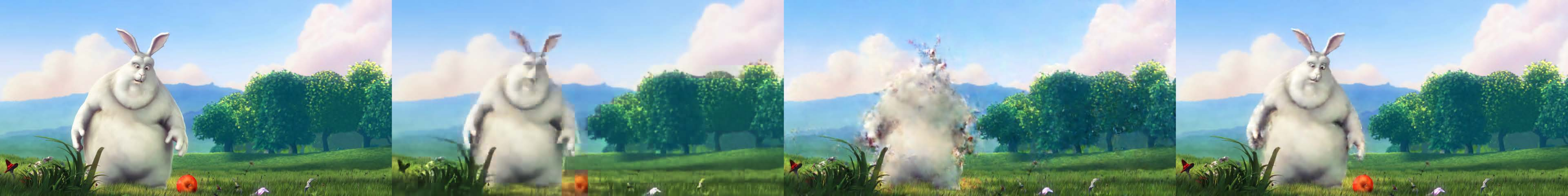}
    \includegraphics[width=.98\linewidth]{figures/vis/unseen_pred_150_gt_ae_nerv_gnerv_23.6_9.21_31.56_full.pdf}

    \caption{Video compression results for unseen frames from Bunny. CNeRV results are similar for unseen as for seen, while autoencoders is still significantly blurrier than CNeRV. NeRV suffers from a myriad of failures.}
    \label{fig:unseen-visulization}
\end{figure*}
We show visualization results for seen images in Figure~\ref{fig:seen-visulization} and unseen images in Figure~\ref{fig:unseen-visulization}.
With the same embedding length and similar total size, CNeRV outperforms NeRV with better detail, absence of artifacts, and lack of spillover from previous frames.
Although autoencoder with large embeddings can reach comparable generalization for unseen images, it struggles a lot for reconstruction of seen images.
But, visual differences still exist between CNeRV and ground truth, and future work can focus on mitigating these issues.
We provide more visualization results in a separate folder named `cnerv\_visualization', you can easily access them through the `index.html'.

\subsection{NeRV Generalization Results}
We provide more NeRV results on shuffled/sequential video frames in Table~\ref{tab:appendix-shuffle-data}, with different number of training images on `Big Buck Bunny'.

\begin{table}[t!]
    \centering
    \caption{NeRV performance with shuffled frame index. It shows that the input embedding of frame index does not provide any meaningful information since shuffling the data does not impact the final performance.}
    \label{tab:appendix-shuffle-data}    
    \vspace{0.5em}
    \begin{tabular}{@{}lccccc}
    \toprule
     &  \multirow{2}{*}{\makecell{Dataset \\  Size}} & \multicolumn{2}{c}{PSNR} & \multicolumn{2}{c}{MS-SSIM} \\
     &  & Seen $\uparrow$ & Unseen $\uparrow$ & Seen$\uparrow$ & Unseen $\uparrow$ \\
     \midrule
    Sequential & 1k & 34.6 & 12.57 & 0.9489 & 0.3495 \\
    Shuffle & 1k & 35.22 & 13.15 & 0.9554 & 0.3905 \\
     \midrule
    Sequential & 2k & 33.53 & 16.46 & 0.919 & 0.4933 \\
    Shuffle & 2k & 33.47 & 16.36 & 0.9193 & 0.4841 \\
     \midrule
    Sequential & 4k & 32.78 & 20.68 & 0.9077 & 0.6994 \\
    Shuffle & 4k & 32.45 & 20.24 & 0.9073 & 0.6859 \\
    \bottomrule
    \end{tabular}
\end{table}

\subsection{Embedding interpolation}
\begin{table}[h!]
\centering
\caption{\textbf{Interpolation} results on different datasets. We show PSNR of unseen images with pixel interpolation and embedding interpolation. Besides, we also show results of ground truth embedding}
\label{tab:interpolate}
\begin{tabular}{@{}c|ccc@{}}
\toprule
 Dataset & \makecell{Pixel \\ Interpolation} & \makecell{GT \\ Embedding}  & \makecell{Embedding \\ Interpolation} \\
 \midrule
UVG & 30.14 & 28.76 & 28.88 \\
MCL & 28.05 & 26.85 & 26.33 \\
Bunny & 27.64 & 26.98 & 24.94 \\
\bottomrule
\end{tabular}
\end{table}

We first provide interpolation results in pixel space and embedding space, together with ground truth embedding results, as shown in Table~\ref{tab:interpolate}. Note that our interpolated embedding shows comparable PSNR with ground truth embedding on unseen images, which clearly demonstrates the internal generalization of our content-adaptive embedding.

\subsection{Main Results with MS-SSIM}

We also provide a detailed main results, with MS-SSIM, in Table ~\ref{tab:appendix-general-dataset}, ~\ref{tab:appendix-general-model-size}, ~\ref{tab:appendix-general-video-resolution}, ~\ref{tab:appendix-general-data-size}, and ~\ref{tab:appendix-general-image-dataset}.

\begin{table*}[h!]
\centering
    
\caption{Results on different \textbf{video datasets}}
\label{tab:appendix-general-dataset}
\vspace{0.5em}
\resizebox{.75\linewidth}{!} {
\begin{tabular}{@{}lccc|ccc|ccc@{}}
\toprule
\makecell{Method} & \makecell{Dataset} & \makecell{Embed \\ Length} & \makecell{Total \\ Size} & \makecell{ \\ Seen} & \makecell{PSNR \\ Unseen $\uparrow$} & \makecell{ \\ Gap $\downarrow$} & \makecell{ \\ Seen} & \makecell{MS-SSIM \\ Unseen $\uparrow$} & \makecell{ \\ Gap $\downarrow$} \\
\midrule
NeRV & UVG &480 &64M & 36.05 & 23.66 & 12.39 & 0.9823 & 0.7314 & 0.2509 \\
CNeRV & UVG &480 &64M & 35.83 & \textbf{28.76} & \textbf{7.07} & 0.9789 & \textbf{0.8739} & \textbf{0.105} \\
\midrule
NeRV & Bunny &480 &64M & 33.53 & 16.46 & 17.07 & 0.919 & 0.4933 & 0.4257 \\
CNeRV & Bunny &480 &64M & 33.83 & \textbf{26.85} & \textbf{6.98} & 0.9539 & \textbf{0.8242} & \textbf{0.1297} \\
\midrule
NeRV & MCL &480 &64M & 34.83 & 19.44 & 15.39 & 0.9815 & 0.5824 & 0.3991 \\
CNeRV & MCL &480 &64M & 34.67 & \textbf{26.98} &\textbf{ 7.69} & 0.978 & \textbf{0.8229} & \textbf{0.1551} \\
\bottomrule
\end{tabular}
}
\vspace{0.5em}

\caption{Results on different \textbf{model sizes} }
\label{tab:appendix-general-model-size}
\resizebox{.75\linewidth}{!} {
\begin{tabular}{@{}lccc|ccc|ccc@{}}
\toprule
\makecell{Method} & \makecell{Model \\ Size} & \makecell{Embed \\ Length} & \makecell{Total \\ Size} & \makecell{ \\ Seen} & \makecell{PSNR \\ Unseen $\uparrow$} & \makecell{ \\ Gap $\downarrow$} & \makecell{ \\ Seen} & \makecell{MS-SSIM \\ Unseen $\uparrow$} & \makecell{ \\ Gap $\downarrow$} \\
\midrule
NeRV & Small &480 &32M & 31 & 16.72 & 14.28 & 0.8977 & 0.5289 & 0.3688 \\
CNeRV & Small &480 &32M & 31.33 & \textbf{26.41} & \textbf{4.92} & 0.911 & \textbf{0.8198} & \textbf{0.0912} \\
\midrule
NeRV & Medium &480 &64M & 33.53 & 16.46 & 17.07 & 0.919 & 0.4933 & 0.4257 \\
CNeRV & Medium &480 &64M & 33.83 & \textbf{26.85} & \textbf{6.98} & 0.9539 & \textbf{0.8242} & \textbf{0.1297} \\
\midrule
NeRV & Large &480 &97M & 35.32 & 16.04 & 19.28 & 0.9485 & 0.4737 & 0.4748 \\
CNeRV & Large &480 &97M & 35.5 & \textbf{27.08} & \textbf{8.42} & 0.9682 & \textbf{0.8235} & \textbf{0.1447} \\
\bottomrule
\end{tabular}
}
\vspace{0.5em}

\caption{Results on different \textbf{video resolutions}}
\label{tab:appendix-general-video-resolution}
\resizebox{.75\linewidth}{!} {
\begin{tabular}{@{}lccc|ccc|ccc@{}}
\toprule
\makecell{Method} & \makecell{Video \\ Resolution} & \makecell{Embed \\ Length} & \makecell{Total \\ Size} & \makecell{ \\ Seen} & \makecell{PSNR \\ Unseen $\uparrow$} & \makecell{ \\ Gap $\downarrow$} & \makecell{ \\ Seen} & \makecell{MS-SSIM \\ Unseen $\uparrow$} & \makecell{ \\ Gap $\downarrow$} \\
\midrule
NeRV & 240*480 & 480 & 60M & 37.14 & 16.9 & 20.24 & 0.9929 & 0.5142 & 0.4787 \\
CNeRV & 240*480 & 480 & 60M & 36.99 & \textbf{27.97} & \textbf{9.02} & 0.9923 & \textbf{0.8532} & \textbf{0.1391} \\
\midrule
NeRV & 480*960 & 480 & 64M & 33.53 & 16.46 & 17.07 & 0.919 & 0.4933 & 0.4257 \\
CNeRV & 480*960 & 480 & 64M & 33.83 & \textbf{26.85} & \textbf{6.98} & 0.9539 & \textbf{0.8242} & \textbf{0.1297} \\
\midrule
NeRV & 960*1920 & 480 & 67M & 32.06 & 16.06 & 16 & 0.902 & 0.5496 & 0.3524 \\
CNeRV & 960*1920 & 480 & 67M & 32.4 & \textbf{26.15} & \textbf{6.25} & 0.9057 & \textbf{0.8118} & \textbf{0.0939} \\
\bottomrule
\end{tabular}
}

\vspace{0.5em}
\caption{Results on different \textbf{training data size}}
\label{tab:appendix-general-data-size}
\resizebox{.75\linewidth}{!} {
\begin{tabular}{@{}lccc|ccc|ccc@{}}
\toprule
\makecell{Method} & \makecell{Training \\ Images} & \makecell{Embed \\ Length} & \makecell{Total \\ Size} & \makecell{ \\ Seen} & \makecell{PSNR \\ Unseen $\uparrow$} & \makecell{ \\ Gap $\downarrow$} & \makecell{ \\ Seen} & \makecell{MS-SSIM \\ Unseen $\uparrow$} & \makecell{ \\ Gap $\downarrow$} \\
\midrule
NeRV & ~1k & 480 & 64M & 34.6 & 12.57 & 22.03 & 0.9489 & 0.3495 & 0.5994 \\
CNeRV & ~1k & 480 & 64M & 34.78 & \textbf{26.41} & \textbf{8.37} & 0.9754 & \textbf{0.813} & \textbf{0.1624} \\
\midrule
NeRV & ~2k & 480 & 64M & 33.53 & 16.46 & 17.07 & 0.919 & 0.4933 & 0.4257 \\
CNeRV & ~2k & 480 & 64M & 33.83 &\textbf{ 26.85} & \textbf{6.98} & 0.9539 & \textbf{0.8242} &\textbf{ 0.1297} \\
\midrule
NeRV & ~4k & 480 & 64M & 32.78 & 20.68 & 12.1 & 0.9077 & 0.6994 & 0.2083 \\
CNeRV & ~4k & 480 & 64M & 32.94 & \textbf{27.75} & \textbf{5.19} & 0.9274 & \textbf{0.8396} &\textbf{ 0.0878} \\
\bottomrule
\end{tabular}
}

\vspace{0.5em}
\caption{Results on different \textbf{image datasets}}
\label{tab:appendix-general-image-dataset}
\resizebox{.75\linewidth}{!} {
\begin{tabular}{@{}lccc|ccc|ccc@{}}
\toprule
\makecell{Method} & \makecell{Dataset} & \makecell{Embed \\ Length} & \makecell{Total \\ Size} & \makecell{ \\ Seen} & \makecell{PSNR \\ Unseen $\uparrow$} & \makecell{ \\ Gap $\downarrow$} & \makecell{ \\ Seen} & \makecell{MS-SSIM \\ Unseen $\uparrow$} & \makecell{ \\ Gap $\downarrow$} \\
\midrule
NeRV & Celeb &240 &33M & 27.44 & 11.27 & 16.17 & 0.9548 & 0.4397 & 0.5151 \\
CNeRV & Celeb &240 &33M & 27.42 & \textbf{21.34} & \textbf{6.08} & 0.9536 & \textbf{0.7879} & \textbf{0.1657} \\
\midrule
NeRV & Flower &240 &35M & 27 & 11.29 & 15.71 & 0.9028 & 0.2538 & 0.649 \\
CNeRV & Flower &240 &36M & 27.04 & \textbf{18.54} & \textbf{8.5} & 0.91 & \textbf{0.5491 }& \textbf{0.3609} \\
\bottomrule
\end{tabular}
}
\end{table*}

\subsection{Embedding Analysis}

We also perform 3 measurements to compare the embeddings produced by CNeRV to prior work.
We compute uniformity to examine the distribution of each set of embeddings on the hypersphere~\citep{wang2020understanding}.

\begin{equation}\label{eq:uniformity}
    U = \text{log} \mathop{\mathbb{E}}_{x,y \sim p_\text{data}}\left[e^{-t||(||f(x)||_2-||f(y)||_2)||_2^2}\right]
\end{equation}

With the same metric, we compute distances between neighbors for each method, to measure the extent to which each method encodes similar representations for neighboring frames. 
We also compute normalized distance, which is the neighbor distance (average distance between neighboring embeddings) divided by the uniformity (average distance between all embeddings).
This normalized distance gives a more accurate reflection of the extent to which a given method's embeddings reflect a semantic connection between neighboring frames.

We compute linear centered kernel alignment (CKA)~\citep{JMLR:v13:cortes12a} to compare the similarity between embeddings of different methods in a pairwise fashion, as introduced by \citep{pmlr-v97-kornblith19a}.
To compute this, we first obtain the matrices containing the embeddings for two different methods, such as NeRV and CNeRV, which we represent $X$ and $Y$. 
We then compute the Gram matrices of the embedding matrices: $K$ = $X X^T$, $L$ = $Y Y^T$. The CKA value is given by the normalized Hilbert-Schmidt Independence Criterion (HSIC)~\citep{gretton2007kernel} in Equation~\ref{eq:cka}.

\begin{equation}\label{eq:cka}
    \text{CKA}(K,L) = \frac{\text{HSIC}(K,L)}{\sqrt{\text{HSIC}(K,K) \text{HSIC}(L,L)}}
\end{equation}

\begin{table}[t!]
\centering
\caption{\textbf{Embedding distribution on hypersphere.}
CNeRV embeddings tend to cluster very tightly together, with low uniformity compared to other methods.
CNeRV neighbor frame embeddings are still reasonably close together, in stark contrast to NeRV.}
\vspace{0.5em}
\label{tab:uniformity}
\begin{tabular}{@{}l|ccc|c|c@{}}
\toprule
\multirow{2}{*}{Methods} & \multicolumn{3}{c}{Uniformity} & \multicolumn{1}{c}{Neighbor} & \multicolumn{1}{c}{Normalized} \\
& Seen $\downarrow$ & Unseen $\downarrow$ & All $\downarrow$ & Distance $\downarrow$ & Distance $\downarrow$ \\
\toprule
FFN~\citep{tancik2020fourier} & 1.18 & 1.17 & 1.18 & 0.09 & 0.07 \\
NeRF~\citep{mildenhall2020nerf} & 1.46 & 1.46 & 1.46 & 0.10 & 0.07 \\
MLF~\citep{mehta2021modulated} & 1.39 & 1.38 & 1.38 & 0.14 & 0.10 \\
ConvAE & 2.00 & 1.97 & 1.99 & 0.14 & \textbf{0.07} \\
ConvVAE & 2.71 & 2.71 & 2.71 & 0.21 & 0.08 \\
NeRV~\citep{chen2021nerv} & 3.97 & 3.96 & 3.97 & 3.60 & 0.91 \\
CNeRV & \textbf{0.23} & \textbf{0.23 } & \textbf{0.23} & \textbf{0.03} & 0.15 \\
\bottomrule
\end{tabular}
\end{table}

\begin{figure}[t!]
    \centering
    \includegraphics[width=.5\linewidth]{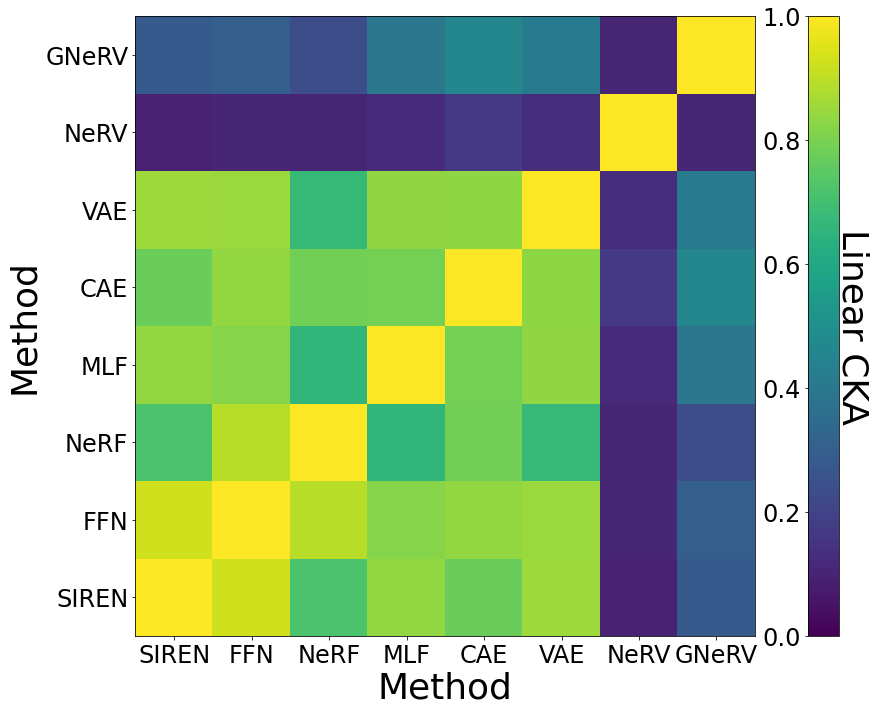}
    \caption{\textbf{Embedding similarity} for methods in Table~\ref{tab:autoencoder}. Embeddings of NeRV and CNeRV are quite different from other methods. }
    \label{fig:combined_cka}
\end{figure}

As Table~\ref{tab:uniformity} shows, while CNeRV's content adaptive embeddings are more tightly clustered than the embeddings for the other methods, CNeRV's neighboring embeddings are still relatively close to each other.
CNeRV thus utilizes very little of the embedding space to generate meaningful and generalizable embeddings, and this is
also consistent with embedding quantization results as in Figure~\ref{fig:embed-quant}.

Figure~\ref{fig:combined_cka} further reflects the difference between CNeRV embeddings and those of other methods.
Whereas the convolutional autoencoders and the pixel-wise neural representations generate somewhat similar embeddings, CNeRV is quite unique by contrast.
This shows how radically different our method is from those already in the literature, in spite of its comparable performance and desirable properties.

\end{document}